\documentclass[letterpaper, 10 pt, conference]{ieeeconf}
\IEEEoverridecommandlockouts
\usepackage{amssymb}
\usepackage{amsmath}
\usepackage{threeparttable}
\usepackage{booktabs}
\usepackage{algorithm}
\usepackage{algorithmicx}
\usepackage{algpseudocode}
\usepackage{xcolor}
\usepackage{graphicx}
\usepackage{subcaption}
\usepackage{multirow}
\usepackage{cite}    
\makeatletter
\let\NAT@parse\undefined
\makeatother
\usepackage{hyperref}
\hypersetup{
hypertex=true,
colorlinks=true,
linkcolor=blue,
anchorcolor=blue,
citecolor=blue
}
\usepackage{xcolor}
\usepackage[table]{xcolor}
\usepackage{pifont}
\usepackage{multirow}
\usepackage{amsfonts}

\begin{document}

\title{\LARGE \bf
GeoSurDepth: Harnessing Foundation Model for Spatial Geometry Consistency-Oriented Self-Supervised Surround-View Depth Estimation}

\author{Weimin Liu$^{1}$, Wenjun Wang$^{1*}$, Joshua H. Meng$^{2}$
\thanks{$^{*}$Corresponding author: Wenjun Wang.}
\thanks{Weimin Liu and Wenjun Wang are with the $^1$State Key Laboratory of Intelligent Green Vehicle and Mobility, School of Vehicle and Mobility, Tsinghua University, Beijing 100084, China (e-mail: lwm23@mails.tsinghua.edu.cn; wangxiaowenjun@tsinghua.edu.cn). Joshua H. Meng is with $^2$California PATH, University of California, Berkeley, CA, United States (e-mail: hdmeng@berkeley.edu).}
}

\maketitle

\begin{abstract}
Accurate surround-view depth estimation provides a competitive alternative to laser-based sensors and is essential for 3D scene understanding in autonomous driving. While empirical studies have proposed various approaches that primarily focus on enforcing cross-view constraints at photometric level, few explicitly exploit the rich geometric structure inherent in both monocular and surround-view setting. In this work, we propose \textbf{GeoSurDepth}, a framework that leverages geometry consistency as the primary cue for surround-view depth estimation. Concretely, we utilize vision foundation models as pseudo geometry priors and feature representation enhancement tool to guide the network to maintain surface normal consistency in spatial 3D space and regularize object- and texture-consistent depth estimation in 2D. In addition, we introduce a novel view synthesis pipeline where 2D-3D lifting is achieved with dense depth reconstructed via spatial warping, encouraging additional photometric supervision across temporal and spatial contexts, and compensating for the limitations of target-view image reconstruction. Finally, a newly-proposed adaptive joint motion learning strategy enables the network to adaptively emphasize informative spatial geometry cues for improved motion reasoning. Extensive experiments on KITTI, DDAD and nuScenes demonstrate that GeoSurDepth achieves SoTA performance, validating the effectiveness of our approach. Our framework highlights the importance of exploiting geometry coherence and consistency for robust self-supervised depth estimation.
\end{abstract}

\section{Introduction}
\par Depth estimation is a fundamental task for 3D scene understanding in autonomous driving. In recent years, self-supervised monocular depth estimation has emerged as a promising approach for 3D perception, eliminating the need for dense groundtruth annotations and making vision-based solutions attractive for large-scale, low-cost deployment \cite{zhou2017unsupervised}. By leveraging photometric reconstruction between consecutive frames or stereo pairs \cite{yang2021self}\cite{poggi2021synergies}, these methods can learn depth directly from raw image sequences. Classical self-supervised approaches, particularly those based on monocular video, typically enforce photometric and smoothness constraints to regularize depth estimations through structure-from-motion (SfM). While effective in single-view scenarios, these methods often suffer from scale ambiguity, temporal inconsistency, and limited geometric reasoning, especially in complex scenes with dynamic objects or low texture regions.
\begin{figure}[t]
    \centering
    \includegraphics[width=1\linewidth]{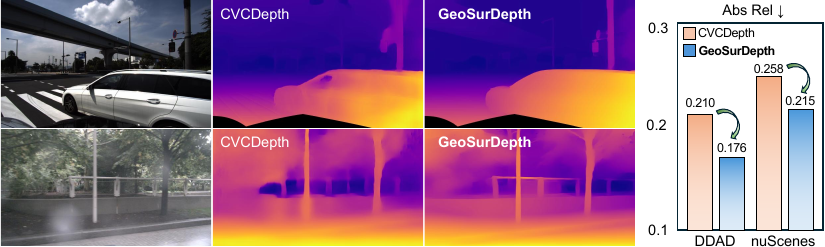}
    \caption{Comparison of depth estimation performance between the proposed method \textbf{GeoSurDepth} and previous method CVCDepth.}
    \label{fig::abstract}
\end{figure}
\par Recently, surround-view depth estimation has received growing attention in autonomous driving and robotics, where multiple cameras collectively capture a $360^\circ$ field of view (FoV) \cite{fsm}. Accurate depth estimation in this setting is crucial for robust scene understanding, obstacle avoidance, and multi-camera fusion. However, extending self-supervised depth estimation to surround-view setups introduces additional challenges, such as cross-view consistency and spatial alignment across cameras. Although several empirical studies have begun addressing these challenges, existing methods still fall short of fully exploiting the rich geometric relationships present in overlapping views, and often fail to properly disentangle photometry, motion, and spatial geometry for consistent multi-camera depth estimation.
\par In this work, we propose \textbf{GeoSurDepth}, a framework designed to address the challenges of geometry consistency in surround-view depth estimation through simple yet effective and intuitive strategies. Our key contributions are summarized as follows: (1) We leverage the powerful vision foundation model DepthAnything (DA) V2 as an indirect pseudo-prior for spatial geometry guidance in surround-view settings. This facilitates accurate and edge-aware self-supervised depth estimation by enforcing 3D surface normal consistency and regularizing object- and texture-level consistency in 2D. Furthermore, we introduce a cross-modal attention module based on CLIP within the depth network to enhance geometric and semantic feature representation. (2) We propose a novel image synthesis approach, where dense depth reconstructed via spatial warping is utilized to achieve 2D-3D lifting, enabling photometric supervision across temporal, spatial, and spatial-temporal domains. This provides a complementary supervision signal, compensating for the limitations of image reconstruction using depth estimated in the target view only. (3) An adaptive joint motion learning strategy is introduced to enhance the network’s interpretability in emphasizing informative camera views for motion cues and learning.
\par In general, in this work, we aim to fully exploit geometry consistency as priors or cues to facilitate depth estimation by tailoring loss function with geometric priors and features, adapting geometry-driven motion learning and enhancing feature representation from geometric perspective.

\section{Related Works}
\par FSM \cite{fsm} is the first work to introduce self-supervised depth estimation to the surround-view setting, aiming to achieve $360^\circ$ dense depth perception. By additionally incorporating photometric reconstruction losses in spatial and spatial-temporal contexts, together with a multi-camera pose consistency constraint, FSM enables metric depth estimation by explicitly exploiting spatial relationship across views. VFDepth \cite{vfdepth} adopts a unified volumetric feature fusion strategy, enabling depth estimation from arbitrary viewpoints. In addition, it proposes a canonical motion estimation strategy that provides a global constraint for the surround-view system and derives per-camera motion via extrinsics-based motion distribution. To further constrain motion estimation and enhance cross-view interaction, SurroundDepth \cite{surrounddepth} estimates joint motion employs a Cross-View Transformer to enrich multi-view feature representations. Subsequently,  MCDP \cite{mcdp} leverages the output of a pre-trained DepthAnything V1 model \cite{depthanythingv1} as pseudo-depth for conditional denoising learning. By iteratively refining cross-view features as conditional inputs, MCDP further improves depth estimation performance. Despite these advances, existing methods primarily focus on pose consistency, feature fusion, or depth refinement, while the explicit geometric consistency and constraints of depth or motion estimation across views remains underexplored, limiting their ability to fully exploit the structural relationships inherent in surround-view camera arrays.
\section{Method}
\label{sec::method}
\subsection{Problem Formulation}
\begin{figure*}
    \centering
    \includegraphics[width=0.95\linewidth]{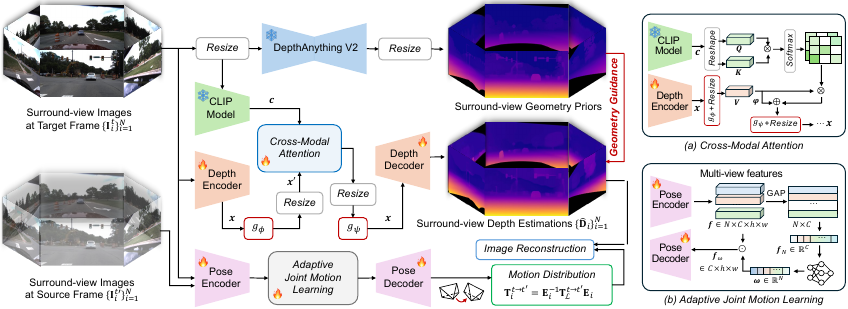}
    \caption{\textbf{Network architecture of GeoSurDepth.} Outputs of DA serve as surround-view geometry priors. Surround-view images at the target frame are first resized to $(518,518)$ before being fed into DA, and the output are interpolated back to original resolution. (a) Cross-modal attention mechanism: For CLIP model, input images are resized to $(214,214)$ for token extraction; (b) Adaptive joint motion learning.}
    \label{fig::overview}
\end{figure*}
\par We formulate surround-view depth estimation in a self-supervised manner under the conventional SfM paradigm, where dense depth and ego-motion are jointly learned from multi-camera image sequences. An overview of the proposed architecture is shown in Fig.\ref{fig::overview}. The framework consists of  trainable depth and pose network, together with frozen vision foundation models including DepthAnything V2 \cite{depthanythingv2} and CLIP \cite{clip} model. 
\par For network training, given surround-view images $\{\mathbf{I}_i^t\}_{i=1}^N$ captured by $N$ cameras at time $t$,  encoder of depth network extracts multi-view features, which are jointly enhanced by fusing CLIP outputs via a cross-modal attention mechanism to improve geometric-semantic coherence. The enhanced features are then passed to depth decoder to produce surround-view depth estimates $\{\hat{\mathbf{D}}_i^t\}_{i=1}^N$, which would be subsequently used for 3D reconstruction and view synthesis within the SfM-based framework. To provide explicit geometric guidance, images at target time are also fed into DA, whose outputs serve as pseudo geometry priors that guide depth network toward geometry-consistent estimations.
\par For motion estimation, the pose network takes temporally adjacent surround-view image pairs $\{(\mathbf{I}_i^t,\mathbf{I}_i^{t'})\}_{i=1}^N$ as input and estimates the corresponding relative camera motions $\{\hat{\mathbf{T}}_i^{t\rightarrow t'}\in \text{SE}(3)\}_{i=1}^N$. During this process, features extracted by pose encoder are processed by the proposed adaptive joint motion learning module, which emphasizes informative camera views before decoding joint ego-motion by pose decoder and distributing motion via calibrated extrinsics. 
\par The estimated depth and pose are jointly used to warp images across views and time, forming the basis for photometric and geometric self-supervision. The entire framework is trained in a fully self-supervised manner, without using groundtruth depth or any pseudo depth for direct supervision. Both pose network and DA are only employed during training and are discarded at inference time. The proposed modules and loss formulations are detailed in the following sections.

\subsection{Spatial Geometry Priors-Guided Self-Supervised Training}
\par\textbf{Photometric loss.} Photometric loss constitutes basic component of self-supervised depth estimation, which calculates the reconstruction error between the target image and synthesized image with not only temporal context, but also spatial and spatial-temporal contexts \cite{fsm} to realize metric estimation. The overall pixel-wise warping operations for image reconstruction are defined as follows, 
\begin{equation}
    \mathbf{p}_{ij}^{t\rightarrow t'}=\mathbf{\Pi}_{ij}^{t\rightarrow t'}\mathbf{p}_i^t,~\tilde{\mathbf{I}}_{ij}^{t\rightarrow t'}(\mathbf{p})=\left<\mathbf{I}_j^{t'}\right>_{\mathbf{p}_{ij}^{t\rightarrow t'}},
    \label{eq::project}
\end{equation}
\begin{equation}
    \mathbf{\Pi}_{ij}^{t\rightarrow t'}=\mathbf{K}_j\mathbf{X}_{ij}^{t\rightarrow t'}\hat{\mathbf{D}}_i\mathbf{K}_i^{-1},
    \label{eq::transform}
\end{equation}
\begin{equation}
    \mathbf{X}_{ij}^{t \rightarrow t^{\prime}}= \begin{cases}\hat{\mathbf{T}}_i^{t \rightarrow t^{\prime}}, & \text{temporal context}, \\ \mathbf{E}_j \mathbf{E}_i^{-1}, & \text{spatial context}, \\ \hat{\mathbf{T}}_j^{t \rightarrow t^{\prime}}\mathbf{E}_j \mathbf{E}_i^{-1}, & \text{spatial-temporal context},\end{cases}
    \label{eq::context}
\end{equation}
where $\mathbf{E}$ and $\mathbf{K}$ indicate extrinsics and intrinsics matrices.
\par The reconstruction error is measured with a weighted sum of intensity difference and structure similarity \cite{godard2017unsupervised}\cite{godard2019digging} as follows,
\begin{equation}
    pe(\mathbf{x}_a,\mathbf{x}_b)=(1-\alpha)\Vert\mathbf{x}_a-\mathbf{x}_b\Vert_1+\alpha\frac{1-\text{SSIM}(\mathbf{x}_a,\mathbf{x}_b)}{2},
\end{equation}
where $\alpha$ is the weighting coefficient, and $\alpha=0.85$.
\par For each context used for pixel-warping, its corresponded and overall photometric loss can be formulated as, 
\begin{equation}
    \begin{cases}
    \mathcal{L}_p^\text{T}=\min_{t'}pe(\mathbf{I}_i^t,\tilde{\mathbf{I}}_i^{t'}),&\text{temporal context}, \\ 
    \mathcal{L}_p^\text{S}=pe(\mathbf{I}_i^t,\tilde{\mathbf{I}}_j^{t}),&\text{spatial context}, \\ \mathcal{L}_p^\text{ST}=\min_{t'}pe(\mathbf{I}_i^t,\tilde{\mathbf{I}}_j^{t'}),&\text{spatial-temporal context},\\
    \mathcal{L}_\text{MVRC}=\min_{t'}pe(\tilde{\mathbf{I}}_j^t,\tilde{\mathbf{I}}_j^{t'}),&\text{MVRC},\\
    
    \end{cases}
\end{equation}
\begin{equation}
    \mathcal{L}_p=\lambda_\text{T}\mathcal{L}_p^\text{T}+\lambda_\text{S}\mathcal{L}_p^\text{S}+\lambda_\text{ST}\mathcal{L}_p^\text{ST}+\lambda_\text{MVRC}\mathcal{L}_\text{MVRC},
\end{equation}
where $\lambda\_$ indicates weight coefficient. MVRC implies the multi-view reconstruction consistency loss proposed by CVCDepth \cite{cvcdepth}, which calculates the photometric error between synthesized image generated with spatial and spatial-temporal contexts within overlapping regions.
\par\textbf{Spatial dense depth-based reconstruction consistency (SRC) loss.} Following the modified spatial backward warping strategy proposed in CVCDepth~\cite{cvcdepth}, we reconstruct a spatial dense depth map in overlapping regions by transforming and projecting depth estimates from adjacent views into the target view. This process can be formulated as,
\begin{equation}
    \mathbf{P}_j=\hat{\mathbf{D}}_j(\mathbf{p}_j)\mathbf{K}_j^{-1}\mathbf{p}_j,~\tilde{\mathbf{P}}_j=\mathbf{E}_i\mathbf{E}_j^{-1}\mathbf{P}_j,
    \label{eq::lift}
\end{equation}
\begin{equation}
    \tilde{\mathbf{D}}_j(\mathbf{p})=\left<(\tilde{\mathbf{P}}_j)_z\right>_{\mathbf{p}_{ij}},
\end{equation}
where $\tilde{\mathbf{P}}_j$ denotes a lifted 3D point from camera $j$ in coordinate frame of camera $i$. $(\cdot)_z$ implies $z$ value of a point cloud. 
\par
\begin{figure}
    \centering
    \includegraphics[width=1\linewidth]{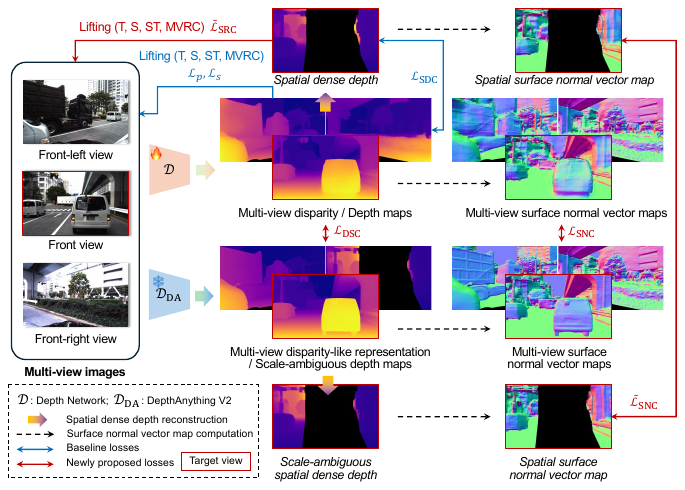}
    \caption{Illustration of spatial geometry priors-guided training.}
    \label{fig::spatial}
\end{figure}

 Based on this modified backward warping of depth map, CVCDepth proposes a spatial dense depth consistency loss to encourage spatial geometry consistency (see Fig.\ref{fig::spatial}). The loss function is formulated as follows. In this work, we also use this loss as part of overall loss function. 
\begin{equation}
    \mathcal{L}_\text{SDC}=\Vert\mathbf{D}_i-\tilde{\mathbf{D}}_j\Vert_1,
\end{equation}
\par Subsequently, we replace $\hat{\mathbf{D}}_i$ with the reconstructed spatial dense depth $\tilde{\mathbf{D}}_j$ in (\ref{eq::transform}) for pixel lifting, and compute photometric losses across temporal, spatial, spatial-temporal, and MVRC contexts, following the same formulation used for directly estimated depth maps in the target view (see Fig.\ref{fig::spatial}). This augmented view synthesis pipeline is however expected to leverage reconstructed spatial dense depth to further enforce cross-view geometric consistency, while compensating for limitations of conventional view synthesis, thereby enabling more robust photometric-level self-supervision. The proposed spatial dense depth-based reconstruction consistency loss can be formulated as,
\begin{equation}
    \tilde{\mathcal{L}}_\text{SRC}=\lambda_\text{T}\tilde{\mathcal{L}}_{p}^\text{T}+\lambda_\text{S}\tilde{\mathcal{L}}_{p}^\text{S}+\lambda_\text{ST}\tilde{\mathcal{L}}_{p}^\text{ST}+\lambda_\text{MVRC}\tilde{\mathcal{L}}_\text{MVRC},
\end{equation}
where $\tilde{\mathcal{L}}$ indicate losses calculated with reconstructed spatial dense depth map reprojected from adjacent views. 
\par\textbf{Geometric prior-guided surface normal consistency (SNC) loss.} Vision foundation models, trained on large-scale datasets with strong generalization capability, can provide valuable prior information for network training. MCDP \cite{mcdp} leverages depth estimations from a pre-trained DepthAnything V1 model as pseudo-depth for conditional denoising learning. VFM-Depth \cite{yu2024vfm} incorporates DINOv2 into the depth encoder to provide universal and stable semantic features. Ninan \textit{et al.} \cite{ninan2025leveraging} used DA V2 for pseudo-supervised training. Instead of directly using depth outputs from foundation models as priors, we compute the corresponding surface normal vector maps and enforce 3D consistency in a scale- and shift-invariant manner. We present the theoretical derivation as follows.
\par Concretely, the output of DA is first normalized to the range $[0,1]$ to mitigate shift effects, yielding a disparity-like representation (see Fig.\ref{fig::spatial}). This representation is then converted into a pseudo scale-ambiguous depth map by applying a clipping range based on the estimated depth (refer to Section 1.1 of Supplementary material for more details). We exploit this to compute surface normal vector map, which is inherently invariant to scale ambiguity. Specifically, we assume $D^\text{true} \approx \gamma D^{\text{DA}}$, where $\gamma$ denotes an unknown scale factor. To obtain surface normal vector, we lift each pixel $\mathbf{p}$ to 3D space as $\mathbf{P} = D^{\text{true}} \mathbf{K}^{-1} \mathbf{p}$. Likewise, we lift two neighboring pixels $\mathbf{p}_1$ and $\mathbf{p}_2$, chosen such that $\cos(\overrightarrow{\mathbf{p}\mathbf{p}_1}, \overrightarrow{\mathbf{p}\mathbf{p}_2})=1$. The surface normal vector at $\mathbf{P}$ can be computed via cross-product as,
\begin{equation}
    \mathbf{n} \propto \mathbf{v}_1 \times \mathbf{v}_2 = \overrightarrow{\mathbf{P}\mathbf{P}_1} \times \overrightarrow{\mathbf{P}\mathbf{P}_2},
    \label{eq::normal}
\end{equation}
where
\begin{equation}
    \mathbf{v}_1 \approx \gamma \left( D_1^{\text{DA}} \mathbf{K}^{-1} \mathbf{p}_1 - D^{\text{DA}} \mathbf{K}^{-1} \mathbf{p} \right):= \gamma \mathbf{a} ,
\end{equation}
\begin{equation}
    \mathbf{v}_2 \approx \gamma \left( D_2^{\text{DA}} \mathbf{K}^{-1} \mathbf{p}_2 - D^{\text{DA}} \mathbf{K}^{-1} \mathbf{p} \right):= \gamma \mathbf{b}.
\end{equation}
\par Substituting the above offset vectors into (\ref{eq::normal}) and normalizing the result to unit length, we obtain,
\begin{equation}
    \mathbf{n} \leftarrow \frac{\mathbf{n}}{\Vert\mathbf{n}\Vert} = \frac{\gamma^2 (\mathbf{a} \times \mathbf{b})}{\gamma^2 \Vert \mathbf{a} \times \mathbf{b} \Vert} = \frac{\mathbf{a} \times \mathbf{b}}{\Vert \mathbf{a} \times \mathbf{b} \Vert},
\end{equation}
which shows that the scale factor $\gamma$ is fully canceled. With this derivation, we hereby constitute our proposed geometric prior-guided surface normal consistency loss as follows.
\par Specifically, we consider eight neighbors of a pixel $\mathbf{p}$ and construct four ordered pixel pairs $\mathcal{P}(\mathbf{p})=\{(\mathbf{p}_{j_0},\mathbf{p}_{j_1})\}_{j=1}^4$, following the strategy proposed in \cite{xue2020toward}. For each pair, the offset vectors relative to $\mathbf{p}$ are mutually perpendicular and arranged in a counterclockwise order. We lift these pixels using depth estimation by the depth network and the pseudo depth of DA, yielding the corresponding 3D point pairs $\mathcal{P}(\mathbf{P})=\{\mathbf{P}_{j_0},\mathbf{P}_{j_1}\}_{j=1}^4$. For each pair, we compute their cross product as described above, from which the surface normal vector map $\mathbf{N}$ is obtained as,
\begin{equation}
    \mathbf{n}_j(\mathbf{P})=\overrightarrow{\mathbf{P}\mathbf{P}_{j_0}}\times\overrightarrow{\mathbf{P}\mathbf{P}_{j_1}}/
    \Vert\overrightarrow{\mathbf{P}\mathbf{P}_{j_0}}\times\overrightarrow{\mathbf{P}\mathbf{P}_{j_1}}\Vert,
\end{equation}
\begin{equation}
    \mathbf{N}(\mathbf{p})=\frac{1}{4}\sum_{j}\texttt{sgn}(\mathbf{n}_0^\top\mathbf{n}_j)\cdot\mathbf{n}_j,
\end{equation}
where $\mathbf{N}\in\mathbb{R}^{3\times1\times H \times W}$. $\mathbf{n}_0$ indicates the surface vector calculated with the first pair in $\mathcal{P}(\mathbf{P})$. To avoid cancellation during averaging, we align the directions of all estimated vectors with $\mathbf{n}_0$ by applying a \texttt{sgn} operation on their inner products. The proposed geometric prior-guided normal consistency loss can thus be formulated as,
\begin{equation}
    \mathcal{L}_\text{SNC}=1-\Vert\hat{\mathbf{N}}^\top\mathbf{N}^\text{DA}\Vert_1,
\end{equation}
where $\hat{\mathbf{N}}$ and $\mathbf{N}^\text{DA}$ indicate surface normal vector map obtained with estimated depth and scale-ambiguous depth map of DA, respectively. Notably, as both $\hat{\mathbf{N}}$ and $\mathbf{N}^\text{DA}$ are normalized to unit length, $\mathcal{L}_\text{SNC}$ could also be regarded as a cosine loss between these vectors. Yet, we use L1 norm to remove the directional ambiguity of surface normals, making the loss invariant to sign flips, which would otherwise penalize orientation discrepancies that have no physical significance when normals are derived from pseudo depth. 
\par Moreover, we augment this loss to $\tilde{\mathcal{L}}_\text{SNC}$ by employing the reconstructed spatial dense depth in previous subsection and form spatial surface normal vector map. We prove that the scale- and shift-invariance still hold for surface normal vector map generated with reconstructed spatial dense depth $\tilde{\mathbf{D}}$ and $\tilde{\mathbf{D}}^\text{DA}$. Detailed deduction of this can be found in Section 1.3 of Supplementary material provided along. 
\par\textbf{Geometric prior-guided disparity smoothness consistency (DSC) loss.} Disparity smoothness loss is commonly used in self-supervised depth estimation task as a regularization term to encourage depth smoothness on inverse depth estimation. Its formulation can be defined as,
\begin{equation}
    \mathcal{L}_s=|\nabla\hat{\boldsymbol{d}}|\cdot \exp(-|\nabla\mathbf{I}|),~\hat{\boldsymbol{d}}:=\hat{\mathbf{D}}^{-1}/\overline{\hat{\mathbf{D}}^{-1}},
\end{equation}
where $\hat{\boldsymbol{d}}$ denotes mean-normalized inverse depth, which emphasizes structural transitions and object boundaries. 
\par To facilitate edge-aware estimation, in prior work of Moon \textit{et al.} \cite{moon2024ground}, object masks introduced to mitigate erroneous depth estimation for dynamic objects by penalizing smoothness transitions between dynamic objects and ground plane, thereby encouraging alignment of the estimated depth of dynamic objects with their contacting ground points. In this work, instead of relying on segmentation cues, we leverage the output of DA not only as a geometric smoothness prior but also as an edge regularizer. This design enhances edge-aware depth estimation, promotes coherent depth transitions, and stabilizes learning in regions where photometric supervision is unreliable. Specifically, we enforce global consistency between the mean-normalized inverse depth gradients of our estimated depth and those derived from DA, which produces depth (or disparity) estimates with clear object-level silhouettes. The resulting loss function is formulated as,
\begin{equation}
    \mathcal{L}_\text{DSC}=\Vert\nabla\hat{\boldsymbol{d}}-\nabla\boldsymbol{d}^\text{DA}\Vert_1,
\end{equation}
which remains scale and shift-invariant due to min-max and mean normalizations.
\par
\begin{figure}[h]
    \centering
    \includegraphics[width=1\linewidth]{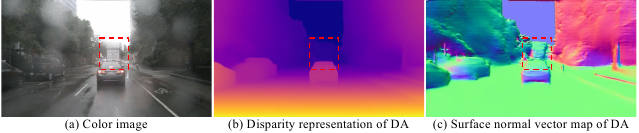}
    \caption{Indication of pseudo geometric priors by DepthAnything V2: Depth of a truck in the front is wrongly estimated.}
    \label{fig::imperfect}
\end{figure}

 In summary, we use surface normal vector map and disparity representation of DA as pseudo geometry priors only. In Fig.\ref{fig::imperfect} we present an imperfect example of DA output.
\par\textbf{Overall loss function.} The overall loss function can be written as,
\begin{equation}
    \mathcal{L}=\mathcal{L}_\text{base}+\sum_{i\in\{\text{SNC},\text{DSC}\}}\omega_i\hat{\mathcal{L}}_i+\mu\sum_{i\in\{\text{SRC},\text{SNC}\}}\kappa_i\tilde{\mathcal{L}}_i,
\end{equation}
\begin{equation}
    \mathcal{L}_\text{base}= \omega_p\hat{\mathcal{L}}_p
 + \omega_s\hat{\mathcal{L}}_s
 + \omega_\text{SDC}\hat{\mathcal{L}}_\text{SDC},
\end{equation}
where we formulate loss function components of CVCDepth as baseline. $\mu$ is weighting coefficient of losses calculated with estimated depth and reconstructed spatial dense depth.
\subsection{Geometric Feature Representation Enhancement}
\par CLIP demonstrates strong capability in capturing high-level semantic representations encoding object- and scene-level priors, which can be complementary to low-level visual features. Motivated by this, we introduce CLIP as an auxiliary semantic encoder to enhance geometry-aware feature representations in the depth estimation pipeline. As illustrated in Fig.\ref{fig::overview}(b), CLIP model is employed to provide high-level semantic cues, which are adaptively fused with image features through a cross-modal attention mechanism, enabling the depth network to leverage semantic consistency for more robust geometric reasoning.
\par Specifically, we extract semantic representations using a frozen CLIP model, yielding token
$\boldsymbol{c}=\texttt{CLIP}(\mathbf{I})\in\mathbb{R}^{N\times T}$,
where $T$ denotes the number of semantic tokens. In parallel, the depth encoder produces multi-scale image features. To facilitate cross-modal attention, we apply a convolutional projection $g_\phi(\cdot)$ and a spatial resizing operation on the last layer output of depth encoder $\boldsymbol{x}\in\mathbb{R}^{N\times C\times h\times w}$,
\begin{equation}
    \boldsymbol{x}'=\texttt{Resize}\big(g_\phi(\boldsymbol{x})\big)\in\mathbb{R}^{N\times C'\times h'w'}.
\end{equation}
\par We then construct a cross-modal attention module, where CLIP token act as the \emph{query} and \emph{key}, while the projected depth features serve as the \emph{value}.
\begin{equation}
    \mathbf{Q}=\texttt{Reshape}(\boldsymbol{c})\in\mathbb{R}^{N\times C'\times L}, ~\mathbf{K}=\mathbf{Q}^\top,~\mathbf{V}=\boldsymbol{x'},
\end{equation}
where $L=h'w'/4$ and $T=C'L$.
\par The final cross-modal attention is computed as,
\begin{equation}
    \boldsymbol{x}' \leftarrow
    \boldsymbol{\varphi}\odot
    \texttt{Softmax}\!\left(
    \frac{\mathbf{Q}\mathbf{K}}{\sqrt{L}}
    \right)\mathbf{V}
    + \boldsymbol{x}',
\end{equation}
where $\boldsymbol{\varphi}\in\mathbb{R}^{C'}$ is a learnable channel-wise scaling factor that adaptively modulates the contribution of attention. We aim to utilize this channel-wise design to stabilize training and allow the network to selectively emphasize meaningful regions for depth estimation. The enhanced features are interpolated back to the original resolution and projected to original channel dimension via convolutional projection $g_\psi(\cdot)$,
\begin{equation}
    \boldsymbol{x} \leftarrow
    g_\psi\big(\texttt{Resize}(\boldsymbol{x}')\big)
    \in\mathbb{R}^{N\times C\times h\times w}.
\end{equation}
\subsection{Adaptive Joint Motion Learning}
\par Pose estimation is a critical component for enabling pixel warping and subsequent view synthesis. Unlike prior approaches that estimate a joint motion in a fixed coordinate frame using features aggregated from all views, such as SurroundDepth and VFDepth, or that assume a fixed camera motion with view-specific features like CVCDepth, we propose an adaptive joint motion learning strategy. Our approach encourages the network to learn and emphasize informative cues for SfM learning, allowing pose estimation to adapt to varying view contributions. 
\par Specifically, we leverage feature maps extracted from all cameras, denoted as $\boldsymbol{f}=\{\boldsymbol{f}_i\}_{i=1}^N$. We first apply spatial average pooling, followed by averaging across the camera dimension to obtain a global feature representation $\bar{\boldsymbol{f}}_N\in\mathbb{R}^C$ that aggregates holistic multi-view information. Instead of uniformly averaging camera features, we introduce a learnable fully-connected network $\xi$ to predict a weight vector $\boldsymbol{\omega}$, enabling adaptive emphasis on informative views for pose estimation. The architecture of the proposed motion learning module is illustrated in Fig.\ref{fig::overview}(a) and is formulated as follows.
\begin{equation}
    \boldsymbol{\omega}=\texttt{softmax}(\xi(\bar{\boldsymbol{f}}_N))\in\mathbb{R}^N,
\end{equation}
\begin{equation}
    \hat{\mathbf{T}}_i^{t\rightarrow t'}=\mathbf{E}_i^{-1}\mathcal{P}_\text{de}\left(\sum_{i}^N{\omega_i\boldsymbol{f}_i}\right)\mathbf{E}_i,
\end{equation}
where $\mathcal{P}_\text{de}$ denotes pose decoder.
\section{Experiments}
\label{sec::experiment}
\subsection{Implementation Details}
\textbf{Dataset.} DDAD \cite{ddad} and nuScenes \cite{nuscenes} provide surround-view imagery captured by six cameras mounted on a vehicle, along with LiDAR point clouds, and are used both training and evaluation in our experiments. For experiments, images are downsampled to $384 \times 640$ for DDAD and $352 \times 640$ for nuScenes.\\
\textbf{Training.} Our networks were implemented in PyTorch and trained on four NVIDIA RTX 4090 GPUs. MonoViT-Small \cite{monovit}, modified and adapted to our surround-view setting, was employed as the depth network. ResNet-18 \cite{resnet} was adopted as pose network following VFDepth \cite{vfdepth}. During training, images from the previous and subsequent frames ($t' \in {t-1, t+1}$) were used as temporal context. We trained the models using the Adam optimizer with $\beta_1=0.9$ and $\beta_2=0.999$, a learning rate of $1\times10^{-4}$, and 30/20 training epochs for DDAD/nuScenes dataset. A batch size of 1, consisting of images from 6 cameras, was used per GPU. Focal normalization \cite{facil2019cam} and the intensity alignment strategy proposed in VFDepth were applied during training (More details are provided in Section 2.1 of Supplementary material).\\
\textbf{Evaluation.} Depth evaluation was conducted up to 200 m for the DDAD dataset and 80 m for the nuScenes dataset. We adopt the depth evaluation metrics proposed in \cite{eigen2014depth} for quantitative comparison unless explicitly label ``scale-ambiguous''. We do not employ horizontal-flip post-processing \cite{godard2019digging} during depth evaluation.

\subsection{Experiment Results}
\par
\begin{figure}[h]
    \centering
    \includegraphics[width=1\linewidth]{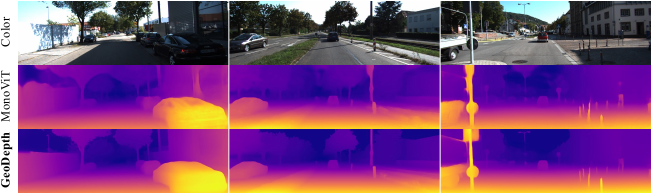}
    \caption{Visualization of depth estimation results on KITTI.}
    \label{fig::kitti_result}
\end{figure}

\textbf{Monocular depth estimation.} We first conduct a simple experiment on monocular depth estimation by altering the network architecture, modifying loss function of MonoViT to incorporate $\mathcal{L}_\text{SNC}$ and $\mathcal{L}_\text{DSC}$, and rename our method as \emph{GeoDepth}. The qualitative results shown in Fig.\ref{fig::kitti_result} demonstrate that our proposed method achieves more accurate, edge- and object-aware depth estimation. The quantitative results reported in Table \ref{table::kitti_short} further show that GeoDepth outperforms MonoViT as well as the SoTA method EDS-Depth \cite{yu2025eds}. Notably, this performance is achieved efficiently and with minimal additional effort, by simply modifying the network architecture and incorporating additional loss terms, without introducing auxiliary modalities such as semantic segmentation, optical flows, or temporal cues. More detailed results are provided in Section 2.3 of the Supplementary material.
\begin{table}[h]
\renewcommand{\arraystretch}{1}
\setlength{\tabcolsep}{2pt}
\centering
\begin{tabular}{c|*{7}{c}}
\toprule
Method & \cellcolor{red!8}{Abs Rel$\downarrow$} & \cellcolor{red!8}{Sq Rel$\downarrow$} & \cellcolor{red!8}{RMSE$\downarrow$} & \cellcolor{red!8}{RMSE$_\text{log}$$\downarrow$} & \cellcolor{cyan!8}{$\delta_1$$\uparrow$} & \cellcolor{cyan!8}{$\delta_2$$\uparrow$} & \cellcolor{cyan!8}{$\delta_3$$\uparrow$}\\
\midrule
MonoViT & 0.099 & 0.708 & 4.372 & 0.175 & 0.900 & 0.967 & 0.984\\
EDS-Depth & \underline{0.095} & \textbf{0.619} & \underline{4.184} & \underline{0.170} & \underline{0.905} & \underline{0.969} & \underline{0.985}\\
\textbf{GeoDepth} & \textbf{0.094} & \underline{0.641} & \textbf{4.175} & \textbf{0.167} & \textbf{0.906} & \textbf{0.970} & \textbf{0.986}\\
\bottomrule
\end{tabular}
\caption{Scale-ambiguous evaluation on KITTI eigen test split. Results best in \textbf{bold}, second best \underline{underlined}).
}
\label{table::kitti_short}
\end{table}
\par
\begin{figure}[h]
    \centering
    \includegraphics[width=1\linewidth]{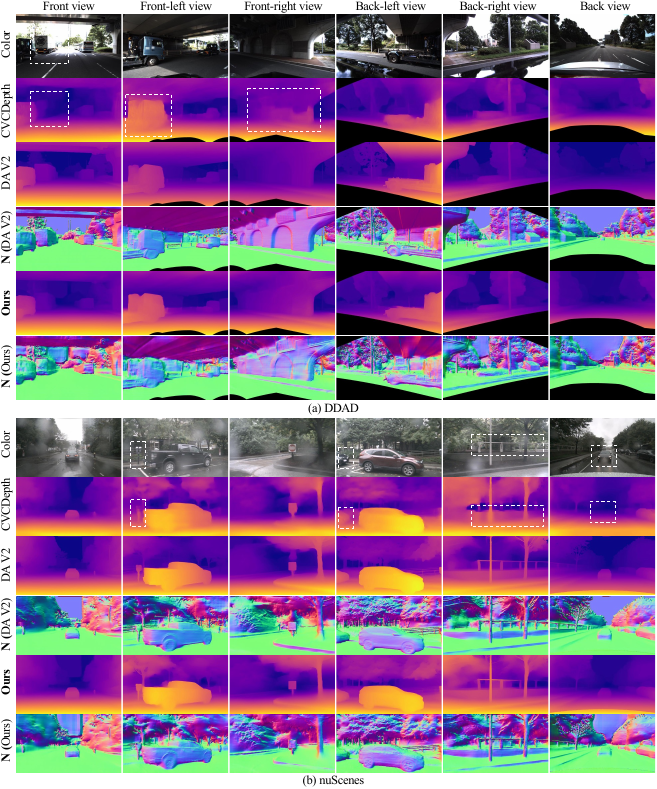}
    \caption{Visualization of depth estimation results on the DDAD and nuScenes datasets. White boxes indicate erroneous estimations.}
    \label{fig::result}
\end{figure}

\textbf{Surround-view depth estimation.} We compare our proposed method with other SoTA approaches. Quantitative evaluations for both metric and scale-ambiguous depth estimation on DDAD and nuScenes datasets are reported in Table \ref{table::result_all}. Our method achieves substantially better or competitive performance and generalization capability compared with other baselines and DepthAnything V2 on both datasets, even when using ResNet-34 as depth network. Qualitative visualizations of surround-view depth estimation results on both datasets are presented in Fig.\ref{fig::result}. As shown, our method produces smooth, edge- and object-aware depth maps, significantly outperforming methods such as CVCDepth. Visualization of more examples as well as point cloud reconstruction on both datasets can be found in Section 2.4 and 2.5 of Supplementary material provided. 
\begin{table*}[t]
\renewcommand{\arraystretch}{1}
\setlength{\tabcolsep}{2.2pt}
\centering
\scalebox{0.97}{
\begin{tabular}{c|*{7}{c}|*{7}{c}}
\toprule
Method & \cellcolor{red!8}{Abs Rel$\downarrow$} & \cellcolor{red!8}{Sq Rel$\downarrow$} & \cellcolor{red!8}{RMSE$\downarrow$} & \cellcolor{red!8}{RMSE$_\text{log}$$\downarrow$} & \cellcolor{cyan!8}{$\delta_1$$\uparrow$} & \cellcolor{cyan!8}{$\delta_2$$\uparrow$} & \cellcolor{cyan!8}{$\delta_3$$\uparrow$} & \cellcolor{red!8}{Abs Rel$\downarrow$} & \cellcolor{red!8}{Sq Rel$\downarrow$} & \cellcolor{red!8}{RMSE$\downarrow$} & \cellcolor{red!8}{RMSE$_\text{log}$$\downarrow$} & \cellcolor{cyan!8}{$\delta_1$$\uparrow$} & \cellcolor{cyan!8}{$\delta_2$$\uparrow$} & \cellcolor{cyan!8}{$\delta_3$$\uparrow$}\\
\midrule
\cellcolor{gray!10}{Scale-aware} & \multicolumn{7}{c}{DDAD} \vline& \multicolumn{7}{c}{nuScenes}\\
\midrule
FSM* \cite{fsm} & 0.228 & 4.409 & 13.433 & 0.342 &  0.687 & 0.870 & 0.932 & 0.319 & 7.534 & 7.860 & 0.362 & 0.716 & 0.874 & 0.931\\
VFDepth \cite{vfdepth} & 0.218 & 3.660 & 13.327 & 0.339 & 0.674 & 0.862 & 0.932 & 0.289 & 5.718 & 7.551 & 0.348 & 0.709 & 0.876 & 0.932\\
SurroundDepth \cite{surrounddepth} & 0.208 & 3.371 & 12.977 & 0.330 & 0.693 & 0.871 & 0.934 & 0.280 & \textbf{4.401} & 7.467 & 0.364 & 0.661 & 0.844 & 0.917 \\
CVCDepth \cite{cvcdepth} & 0.210 & 3.458 & 12.876 & - & 0.704 & - & - & 0.258 & 4.540 & 7.030 & - & 0.756 & - & - \\
SA-FSM \cite{sa-fsm} & 0.187 & 3.093 & \underline{12.578} & 0.311 & \underline{0.731} & \underline{0.891} & \underline{0.945} & 0.272 & 4.706 & 7.391 & 0.355 & 0.689 & 0.868 & 0.929 \\
\textbf{GeoSurDepth} (Res34) & \underline{0.184} & \underline{2.896} & 12.912 & \underline{0.303} & 0.729 & 0.889 & 0.944 & \underline{0.220} & \underline{4.537} & \underline{6.196} & \underline{0.287} & \underline{0.811} & \underline{0.915} & \underline{0.951}\\
\textbf{GeoSurDepth} (proposed) & \textbf{0.176} & \textbf{2.738} & \textbf{11.520} & \textbf{0.280} & \textbf{0.763} & \textbf{0.912} & \textbf{0.957} & \textbf{0.215} & 4.845 & \textbf{6.157} & \textbf{0.282} & \textbf{0.823} & \textbf{0.922} & \textbf{0.954}\\
\midrule

\cellcolor{gray!10}{Scale-ambiguous} & \multicolumn{7}{c}{DDAD} \vline& \multicolumn{7}{c}{nuScenes}\\
\midrule
FSM* \cite{fsm} & 0.219 & 4.161 & 13.163 & 0.327 & 0.703 & 0.880 & 0.940 & 0.301 & 6.180 & 7.892 & 0.366 & 0.729 & 0.876 & 0.933\\
VFDepth \cite{vfdepth} & 0.221 & 3.549 & 13.031 & 0.323 & 0.681 & 0.874 & 0.940 & 0.271 & 4.496 & 7.391 & 0.346 & 0.726 & 0.879 & 0.934\\ 
SurroundDepth \cite{surrounddepth} & 0.200 & 3.392 & 12.270 & 0.301 & 0.740 & 0.894 & 0.947 & 0.245 & 3.067 & 6.835 & 0.321 & 0.719 & 0.878 & 0.935 \\
CVCDepth \cite{cvcdepth} & 0.208 & 3.380 & 12.640 & - & 0.716 & - & - & 0.258 & 4.540 & 7.030 & - & 0.756 & - & - \\
SA-FSM \cite{sa-fsm} & 0.189 & 3.130 & 12.345 & 0.299 & 0.744 & 0.897 & 0.949 & 0.245 & 3.454 & 6.999 & 0.325 & 0.725 & 0.875 & 0.934\\
MCDP \cite{mcdp} & 0.187 & 2.983 & \underline{11.745} & - & \textbf{0.831} & - & - & 0.213 & \textbf{2.858} & 6.346 & - & 0.775 & - & -\\
DepthAnything V2 \cite{depthanythingv2} & 0.181 & 4.395 & 13.816 & \underline{0.288} & 0.768 & \underline{0.905} & \underline{0.952} & 0.269 & 5.361 & 8.757 & 0.343 & 0.707 & 0.863 & 0.925 \\
\textbf{GeoSurDepth} (Res34) & \underline{0.180} & \underline{2.820} & 12.640 & 0.290 & 0.748 & 0.898 & 0.950 & \underline{0.208} & \underline{2.872} & \underline{6.169} & \underline{0.283} & \underline{0.793} & \underline{0.907} & \underline{0.949}\\
\textbf{GeoSurDepth} (proposed) & \textbf{0.167} & \textbf{2.639} & \textbf{11.381} & \textbf{0.268} & \underline{0.786} & \textbf{0.916} & \textbf{0.959} & \textbf{0.197} & 2.952 & \textbf{6.050} & \textbf{0.276} & \textbf{0.810} & \textbf{0.913} & \textbf{0.951}\\

\midrule
\cellcolor{gray!10}{Cross-dataset (Scale-ambiguous)} & \multicolumn{7}{c}{Trained on nuScenes, tested on DDAD} \vline& \multicolumn{7}{c}{Trained on DDAD, tested on nuScenes}\\
\midrule
CVCDepth (Res34) \cite{cvcdepth} & 0.270 & 5.273 & 15.127 & 0.400 & 0.603 & 0.810 & 0.895 & 0.289 & 3.503 & 7.469 & 0.353 & 0.642 & 0.848 & 0.923\\
\textbf{GeoSurDepth} (proposed) & 0.193 & 3.049 & 13.227 & 0.307 & 0.719 & 0.885 & 0.944 & 0.246 & 2.688 & 6.885 & 0.318 & 0.694 & 0.877 & 0.937 \\

\bottomrule
\end{tabular}
}
\caption{Depth evaluation and generalization test results on DDAD and nuScenes datasets (* indicates reproduced results by VFDepth. Generalization tests were conducted using median-scaling due to different depth clipping ranges across two datasets.}
\label{table::result_all}
\end{table*}

\subsection{Ablation Studies}
\par\textbf{Adaptive joint motion learning.} Table \ref{table::ab_pose} shows improved depth estimation performance with proposed method against other baselines or methods for individual camera views and surround-view setting. We attribute this improvement to the learnable weighting of features extracted by the pose encoder, which enables the network to adaptively emphasize informative views and thereby achieve more effective SfM learning.
\begin{table}[h]
\renewcommand{\arraystretch}{1}
\setlength{\tabcolsep}{2.5pt}
\centering
\scalebox{0.95}{
\begin{tabular}{c|*{7}{c}}
\toprule
\multirow{2}{*}{\vspace{-0.5em}Method} & \multicolumn{7}{c}{\cellcolor{red!8}{Abs Rel$\downarrow$}}\\
\cmidrule{2-8}
& \textit{F} & \textit{FL} & \textit{FR} & \textit{BL} & \textit{BR} & \textit{B} & \textit{Avg.}\\
\midrule
Pose consistency & 0.142 & 0.179 & 0.217 & 0.192 & 0.216 & 0.180 & 0.188\\
Joint pose & 0.131 & 0.179 & \underline{0.200} & 0.193 & 0.215 & 0.168 & 0.181\\
Canonical front pose & \underline{0.130} & \underline{0.174} & \textbf{0.199} & \underline{0.188} & 0.212 & \underline{0.162} & \underline{0.177}\\
Front pose & 0.131 & 0.178 & \textbf{0.199} & 0.192 & \underline{0.209} & 0.164 & 0.179\\
\midrule
SurroundDepth & 0.152 & 0.207 & 0.230 & 0.220 & 0.239 & 0.200 & 0.208\\
CVCDepth (Res34) & 0.146 & 0.201 & 0.221 & 0.220 & 0.232 & 0.192 & 0.202\\
\midrule
\textbf{Adaptive joint motion} & \textbf{0.129} & \textbf{0.171} & 0.204 & \textbf{0.187} & \textbf{0.206} & \textbf{0.159} & \textbf{0.176}\\
\bottomrule
\end{tabular}
}
\caption{Ablation study on different motion learning strategies on DDAD dataset (\emph{F}, \emph{B}, \emph{L}, \emph{R} implies front, back, left and right).}
\label{table::ab_pose}
\end{table}

\par
\begin{figure}
    \centering
    \includegraphics[width=1\linewidth]{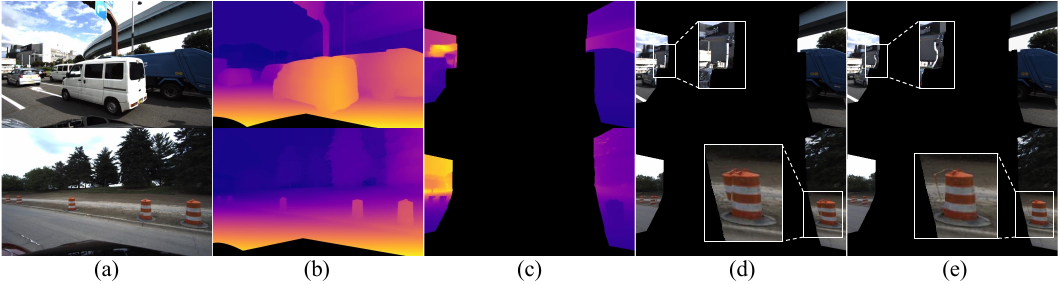}
    \caption{View synthesis example: (a) Color image; (b) Disparity map; (c) Reconstructed spatial dense depth; (d)(e) Spatial warping with estimated depth and reconstructed spatial dense depth.}
    \label{fig::ablation_spatial}
\end{figure}

\textbf{Spatial dense depth-based view synthesis.} In this work, we leverage reconstructed spatial dense depth to further perform view synthesis and enforce surface normal consistency. As shown in Table \ref{table::ab_spatial_depth}, removing either $\tilde{\mathcal{L}}_\text{SRC}$ or $\tilde{\mathcal{L}}_\text{SNC}$ from the overall loss function leads to noticeable performance degradation. Fig.\ref{fig::ablation_spatial} presents a qualitative example of view synthesis using both the estimated depth and the reconstructed spatial dense depth through spatial warping. The zoomed-in regions illustrate the complementary effect of these two depth sources on view synthesis and photometric penalization for 3D lifting.
\begin{table}[h]
\renewcommand{\arraystretch}{1}
\setlength{\tabcolsep}{2pt}
\centering
\scalebox{0.95}{
\begin{tabular}{cc|*{7}{c}}
\toprule
$\tilde{\mathcal{L}}_\text{SRC}$ & $\tilde{\mathcal{L}}_\text{SNC}$ & \cellcolor{red!8}{Abs Rel$\downarrow$} & \cellcolor{red!8}{Sq Rel$\downarrow$} & \cellcolor{red!8}{RMSE$\downarrow$} & \cellcolor{red!8}{RMSE$_\text{log}$$\downarrow$} & \cellcolor{cyan!8}{$\delta_1$$\uparrow$} & \cellcolor{cyan!8}{$\delta_2$$\uparrow$} & \cellcolor{cyan!8}{$\delta_3$$\uparrow$}\\
\midrule
\ding{55} & \ding{55} & 0.183 & 2.746 & \underline{11.557} & \underline{0.281} & 0.754 & \underline{0.909} & \underline{0.956}\\
\ding{51} & \ding{55} & \underline{0.180} & 2.773 & 11.647 & 0.282 & \underline{0.757} & \underline{0.909} & \underline{0.956}\\
\ding{55} & \ding{51} &  0.181 & \textbf{2.734} & 12.014 & 0.287 & 0.747 & 0.904 & 0.954\\
\midrule
\ding{51} & \ding{51} & \textbf{0.176} & \underline{2.738} & \textbf{11.520} & \textbf{0.280} & \textbf{0.763} & \textbf{0.912} & \textbf{0.957}\\
\bottomrule
\end{tabular}
}
\caption{Evaluation of the use of reconstructed spatial dense depth on metric depth estimation of DDAD dataset.}
\label{table::ab_spatial_depth}
\end{table}
\par
\begin{figure}
    \centering
    \includegraphics[width=1\linewidth]{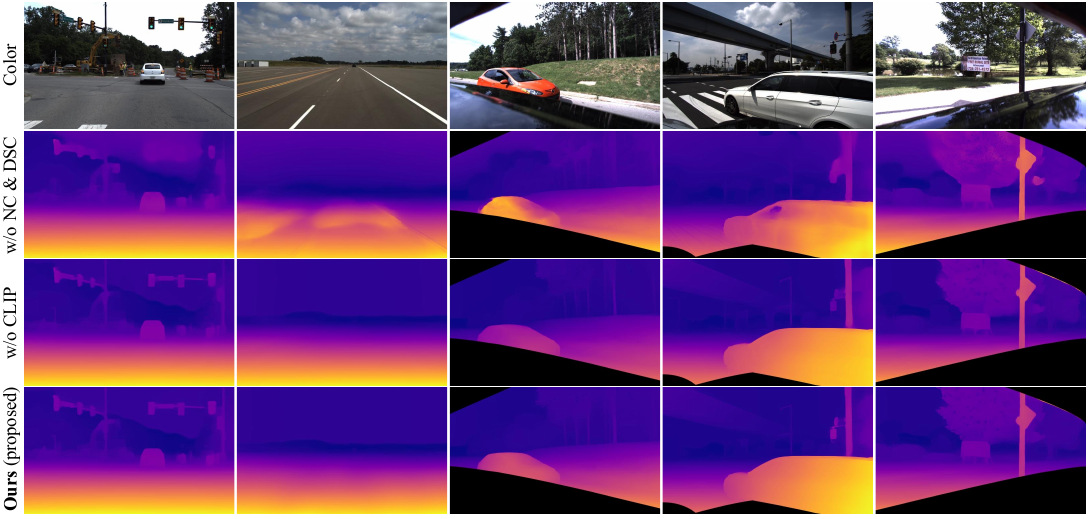}
    \caption{Visualization results of ablation study on geometry guidance by DA and feature representation enhancement by CLIP model.}
    \label{fig::ablation_foundation}
\end{figure}

\textbf{Geometry guidance and enhancement via foundation models.} Results reported in Table~\ref{table::ab_foundation} demonstrate that enforcing 3D geometric accuracy through surface normal consistency, together with regularizing edge-aware depth estimation using 2D disparity gradients, leads to improved model performance. In contrast, removing CLIP from the depth network results in a slight drop in model performance. Fig.\ref{fig::ablation_foundation} provides qualitative comparisons of these variants, illustrating less smooth depth transitions in low-texture regions, increased edge blurring at object boundaries when geometric guidance is absent, and reduced semantic details when geometric feature representation enhancement is removed.
\begin{table}[h]
\renewcommand{\arraystretch}{1}
\setlength{\tabcolsep}{1.9pt}
\centering
\scalebox{0.93}{
\begin{tabular}{c|*{7}{c}}
\toprule
Method & \cellcolor{red!8}{Abs Rel$\downarrow$} & \cellcolor{red!8}{Sq Rel$\downarrow$} & \cellcolor{red!8}{RMSE$\downarrow$} & \cellcolor{red!8}{RMSE$_\text{log}$$\downarrow$} & \cellcolor{cyan!8}{$\delta_1$$\uparrow$} & \cellcolor{cyan!8}{$\delta_2$$\uparrow$} & \cellcolor{cyan!8}{$\delta_3$$\uparrow$}\\
\midrule
w/o SNC\&DSC & 0.190 & 3.054 & 11.959 & 0.299 & 0.742 & 0.899 & 0.949\\
w/o DSC & 0.184 & 3.064 & 11.965 & 0.297 & 0.743 & 0.900 & 0.949\\
w/o SNC &  0.182 & 2.883 & 12.230 & 0.297 & 0.734 & 0.898 & 0.950\\
\midrule
w/o CLIP & \underline{0.179} & \textbf{2.731} & \underline{11.578} & \underline{0.281} & \underline{0.757} & \underline{0.911} & \underline{0.956}\\
\midrule
\textbf{Ours} &  \textbf{0.176} & \underline{2.738} & \textbf{11.520} & \textbf{0.280} & \textbf{0.763} & \textbf{0.912} & \textbf{0.957}\\
\bottomrule
\end{tabular}
}
\caption{Ablation study on geometry guidance by DepthAnything V2 and feature representation enhancement by CLIP model.}
\label{table::ab_foundation}
\end{table}

\section{Conclusion}
\label{sec::conclusion}
In this work, we presented GeoSurDepth, a framework for self-supervised surround-view depth estimation that exploits geometry consistency as primary cue. By integrating vision foundation models as pseudo-geometry priors and for feature enhancement, enforcing 3D surface normal consistency, and regularizing object- and texture-level depth across views, GeoSurDepth achieves accurate and edge-aware depth estimations. A novel view synthesis pipeline provides additional photometric supervision through 2D-3D lifting and multi-contextual reconstruction, while an adaptive joint motion learning strategy enables the network to emphasize informative camera views for improved motion reasoning. Extensive experiments on KITTI, DDAD and nuScenes demonstrate that GeoSurDepth achieves SoTA performance, highlighting the importance of exploiting geometry coherence and consistency for robust self-supervised depth estimation. 

\bibliographystyle{IEEEtran}
\bibliography{reference.bib}

\end{document}


\title{\LARGE \bf
Supplementary material for:\\
GeoSurDepth: Harnessing Foundation Model for Spatial Geometry Consistency-Oriented Self-Supervised Surround-View Depth Estimation}

\author{Weimin Liu$^{1}$, Wenjun Wang$^{1*}$, Joshua H. Meng$^{2}$
}

\maketitle

\section{Method}
\label{sec::method}

\subsection{Pseudo depth of DepthAnything V2}
We denote the inverse depth representation directly output by DepthAnything V2 as $\mathbf{S}^{\text{DA}}$. Due to the inherent scale and shift ambiguity of monocular depth estimation, this representation cannot be directly interpreted as a metric quantity. Instead, it can be formulated as an affine transformation of the true inverse depth,
\begin{equation}
    \mathbf{S}^{\text{DA}} = \alpha \frac{1}{\mathbf{D}^{\text{true}}} + \beta,
    \label{eq::1}
\end{equation}
where $\mathbf{D}^{\text{true}}$ denotes the true depth in the physical world, and $\alpha$ and $\beta$ are unknown scale and offset parameters.
\par To remove this affine ambiguity, we apply min-max normalization to $\mathbf{S}^{\text{DA}}$,
\begin{equation}
    \bar{\mathbf{S}}^\text{DA}
    = \frac{\mathbf{S}^{\text{DA}} - \mathbf{S}^{\text{DA}}_\text{min}}
    {\mathbf{S}^{\text{DA}}_\text{max} - \mathbf{S}^{\text{DA}}_\text{min}}.
    \label{eq::2}
\end{equation}
\par Substituting (\ref{eq::1}) into (\ref{eq::2}) gives
\begin{equation}
    \bar{\mathbf{S}}^\text{DA}=
    \frac{\frac{\alpha}{\mathbf{D}^{\text{true}}} + \beta
    - \left(\frac{\alpha}{\mathbf{D}^{\text{true}}_\text{max}} + \beta\right)}
    {\left(\frac{\alpha}{\mathbf{D}^{\text{true}}_\text{min}} + \beta\right)
    - \left(\frac{\alpha}{\mathbf{D}^{\text{true}}_\text{max}} + \beta\right)} = \frac{\frac{1}{\mathbf{D}^{\text{true}}}
    - \frac{1}{\mathbf{D}^{\text{true}}_\text{max}}}
    {\frac{1}{\mathbf{D}^{\text{true}}_\text{min}}
    - \frac{1}{\mathbf{D}^{\text{true}}_\text{max}}},
\end{equation}
which shows that the normalized representation $\bar{\mathbf{S}}^\text{DA}$ is invariant to the unknown affine parameters $\alpha$ and $\beta$, and depends solely on the relative inverse-depth distribution.
\par We further interpret $\bar{\mathbf{S}}^\text{DA}$ as a normalized disparity-like representation and map it to a predefined target depth range $[\mathbf{D}^{\text{tgt}}_\text{min}, \mathbf{D}^{\text{tgt}}_\text{max}]$. Specifically, we define
\begin{equation}
    \text{disp}_\text{min} = \frac{1}{\mathbf{D}^{\text{tgt}}_\text{max}},~
    \text{disp}_\text{max} = \frac{1}{\mathbf{D}^{\text{tgt}}_\text{min}},
\end{equation}
and recover the estimated depth as,
\begin{equation}
    \begin{aligned}
    \mathbf{D}^{\text{DA}}
    &= \frac{1}{\text{disp}_\text{min}
    + (\text{disp}_\text{max} - \text{disp}_\text{min}) \cdot \bar{\mathbf{S}}^\text{DA}
    } \\
    &= \frac{1}{\frac{1}{\mathbf{D}^{\text{tgt}}_\text{max}}
    + \left(\frac{1}{\mathbf{D}^{\text{tgt}}_\text{min}}
    - \frac{1}{\mathbf{D}^{\text{tgt}}_\text{max}}\right)
    \cdot \bar{\mathbf{S}}^\text{DA}
    } \\
    &= \frac{1}{\frac{1}{\mathbf{D}^{\text{tgt}}_\text{max}}
    + \left(\frac{1}{\mathbf{D}^{\text{tgt}}_\text{min}}
    - \frac{1}{\mathbf{D}^{\text{tgt}}_\text{max}}\right)
    \cdot
    \frac{1/\mathbf{D}^{\text{true}}
    - 1/\mathbf{D}^{\text{true}}_\text{max}}
    {1/\mathbf{D}^{\text{true}}_\text{min}
    - 1/\mathbf{D}^{\text{true}}_\text{max}} }.
    \end{aligned}
\end{equation}
\par
\begin{figure*}
    \centering
    \includegraphics[width=1\linewidth]{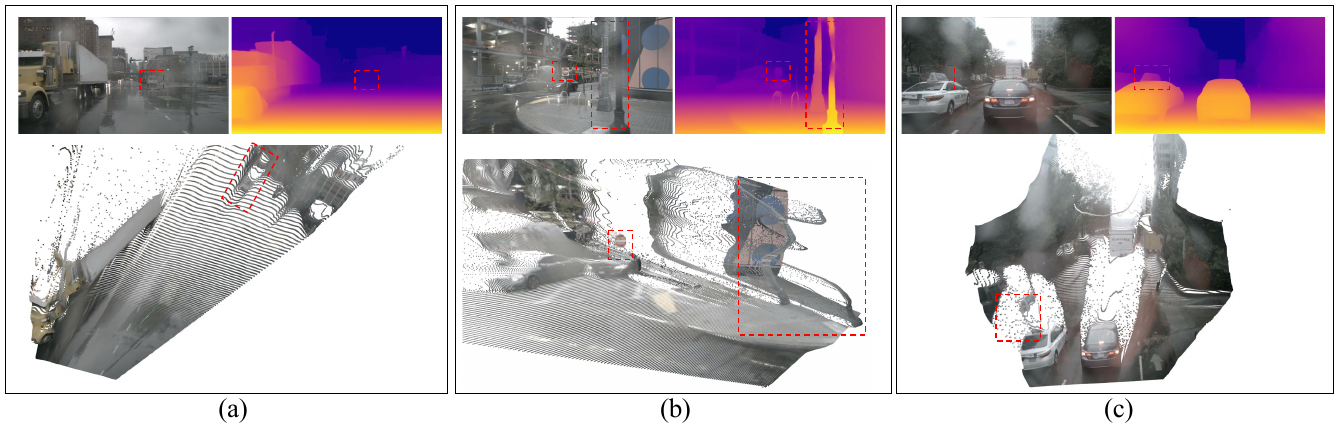}
    \caption{Examples point cloud reconstruction using of pseudo depth of DepthAnything V2 on nuScenes dataset (upper left: RGB color image; upper right: pseudo depth of DepthAnything V2; bottom: reconstructed point cloud): (a) the vehicle ahead is incorrectly projected in 3D; (b) the light pole is skewed in the 3D projection; (c) the overhead taxi sign is incorrectly projected.}
    \label{fig::pseudo_prior}
\end{figure*}

 In practice, both $\mathbf{D}^{\text{true}}_\text{max}$ and $\mathbf{D}^{\text{tgt}}_\text{max}$ are typically large, such that their reciprocals can be approximated as zero. Under this approximation, the above expression simplifies to
\begin{equation}
\mathbf{D}^{\text{DA}}
\approx \frac{1/\mathbf{D}^{\text{true}}_\text{min}
- 1/\mathbf{D}^{\text{true}}_\text{max}}
{1/\mathbf{D}^{\text{tgt}}_\text{min}
- 1/\mathbf{D}^{\text{tgt}}_\text{max}}
\cdot \mathbf{D}^{\text{true}}
= \frac{1}{\gamma} \mathbf{D}^{\text{true}},
\end{equation}
where $\gamma$ is a scale factor that characterizes the proportional relationship between the pseudo depth inferred from DepthAnything V2 and the true depth in the physical world. Based on this property, we further compute the surface normal vector map from the pseudo depth by DA, and reconstruct spatial dense depth through cross-view geometry. Notably, we use the surface normal vector map solely as a pseudo prior, since the scale-ambiguous depth estimated by DepthAnything V2 can only be treated as pseudo labels. We present the point cloud reconstruction effect using this pseudo depth in Fig.\ref{fig::pseudo_prior}. 

\subsection{Spatial dense depth reconstruction methods}
\par In this work, we adopt the modified spatial backward warping strategy proposed in CVCDepth \cite{cvcdepth} to reconstruct spatial dense depth from adjacent views. We further summarize representative depth reprojection and reconstruction strategies employed in prior studies.
\par (1) \textit{Forward warping (FW)} \cite{cvcdepth19}. FW lifts each pixel from source view into 3D space using the estimated depth, and transforms it into target camera coordinate system via extrinsics. The transformed 3D point is then projected onto the target image plane, where its depth value is assigned to corresponding pixel to form warped depth map. FW is geometrically correct. However, it does not define a one-to-one mapping and also results in holes in depth map due to discretization. FW can be formulated as,
\begin{equation}
    \mathbf{P}_j=\hat{\mathbf{D}}_j(\mathbf{p}_j)\mathbf{K}_j^{-1}\mathbf{p}_j,
    \label{eq::lift}
\end{equation}
\begin{equation}
    \tilde{\mathbf{P}}_j=\mathbf{E}_i\mathbf{E}_j^{-1}\mathbf{P}_j,~\tilde{\mathbf{D}}_j(\mathcal{Q}(\mathbf{K}_i\tilde{\mathbf{P}}_j))=(\tilde{\mathbf{P}}_j)_z,
\end{equation}
where $\tilde{\mathbf{P}}_j\in\mathbb{R}^3$ denotes a 3D point reprojected from camera $j$ to the coordinate frame of camera $i$. $(\cdot)_z$ implies $z$ value of a point cloud. $\mathcal{Q}(\cdot)$ denotes discretization operator that maps continuous coordinates to pixel indices.  

\par (2) \textit{Backward warping (BW)}. BW follows the same synthesis procedure as bilinear-sampling-based view synthesis with only source frame of a depth map. It does not provide correct depth reference as same objects have different depth value in different viewpoints. BW can be formulated as follows.
\begin{equation}
    \mathbf{p}_{ij}=\mathbf{\Pi}_{ij}\mathbf{p}_i,~\tilde{\mathbf{D}}_j(\mathbf{p})=\left<\hat{\mathbf{D}}_j\right>_{\mathbf{p}_{ij}}
\end{equation}
\par (3) \textit{Modified backward warping (MBW)}. To deal with issue of BW, CVCDepth \cite{cvcdepth} implements bilinear sampling on source depth map transformed to target view to facilitate geometry correctness. MBW can be formulated as follows.
\begin{equation}
    \tilde{\mathbf{D}}_j(\mathbf{p})=\left<(\tilde{\mathbf{P}}_j)_z\right>_{\mathbf{p}_{ij}}
\end{equation}
\par (4) \textit{Modified forward+backward warping (MFBW)}. MFBW, proposed and used in MonoDiffusion \cite{monodiffusion} and \cite{liu2023self}, constructs depth filtering masks for knowledge distillation from a teacher network. It first obtains a depth map via BW and uses it to lift 2D pixels into 3D space, followed by a 3D-2D projection as in FW to deal with discretization issue. However, the reprojection relies on bilinearly sampled depth values on the source view, which may correspond to interpolated 3D points that do not strictly exist in the physical scene. MFBW can be written as follows,
\begin{equation}
    \mathbf{P}_j^\text{BW}=\hat{\mathbf{D}}_j^\text{BW}(\mathbf{p}_j)\mathbf{K}_j^{-1}\mathbf{p}_j,
\end{equation}
\begin{equation}
    \tilde{\mathbf{P}}_j^\text{BW}=\mathbf{E}_i\mathbf{E}_j^{-1}\mathbf{P}_j,~\tilde{\mathbf{D}}_j(\mathbf{p})=(\tilde{\mathbf{P}}_j^\text{BW})_z.
\end{equation}
\subsection{Proof of scale- and shift-invariance in surface normal vector map computation with reconstructed spatial dense depth of DepthAnything V2}
\par In this work, both $\mathcal{L}_\text{SNC}$ and $\tilde{\mathcal{L}}_\text{SNC}$ are adopted as components of the overall loss function. In the submitted main manuscript, we present a derivation demonstrating the scale invariance of surface normal vector computation when using the scale-ambiguous depth produced by DA. Here, we provide a detailed derivation showing that the surface normal vector map computed from reconstructed spatial dense depth of Depth is also invariant to scale and shift.
\par Specifically, we perform modified spatial backward warping on the pseudo scale-ambiguous depth by DepthAnything V2 as follows.
\begin{equation}
    \mathbf{P}_j=D_j^\text{true}\mathbf{K}^{-1}_j\mathbf{p}_j\approx\gamma D_j^\text{DA}\mathbf{K}_j^{-1}\mathbf{p}_j,
\end{equation}
\begin{equation}
\begin{aligned}
    & \tilde{D}_j^\text{DA}=\left<(\tilde{\mathbf{P}}_j)_z\right>_{\mathbf{p}_{ij}}=\left<(\mathbf{R}_{ij}\mathbf{P}_j+\mathbf{t}_{ij})_z\right>_{\mathbf{p}_{ij}},\\
    & \Rightarrow \tilde{D}_j^\text{DA}=\left<\gamma(\mathbf{R}_{ij}D_j^\text{DA}\mathbf{K}_j^{-1}\mathbf{p}_j)_z+z_{ij}\right>_{\mathbf{p}_{ij}},
\end{aligned}
\label{eq::D_DA}
\end{equation}
where $[\mathbf{R}_{ij}, \mathbf{t}_{ij}]=\mathbf{E}_j^{-1}\mathbf{E}_i$, $z_{ij}=(\mathbf{t}_{ij})_z$, $\mathbf{p}_{ij}=\mathbf{R}_{ij}\mathbf{p}_i+\mathbf{t}_{ij}$, $\left<\cdot\right>$ indicates bilinear sampling. 
\par We simplify above formulation (\ref{eq::D_DA}) as,
\begin{equation}
    \check{D}\approx\check{\gamma}D_j^\text{DA}+\check{z},
\end{equation}
where $\check{\gamma}$ and $\check{z}$ are ambiguous scale and shift, respectively. We use unbolded notation to denote the value at an individual pixel for notational simplicity. For example, $D_j^\text{DA}=\mathbf{D}_j^\text{DA}(\mathbf{p})$. 
\par Similarly, we lift 2D pixels with this depth, yielding,
\begin{equation}
    \mathbf{P}=\check{D}\mathbf{K}^{-1}\mathbf{p}=\check{\gamma}D_j^\text{DA}\mathbf{K}^{-1}\mathbf{p}+\check{z}\mathbf{K}^{-1}\mathbf{p}.
\end{equation}
\par The offset vector can thus be calculated as follows.
\begin{equation}
    \mathbf{v}_1=\check{\gamma}D_{j,1}^\text{DA}\mathbf{K}^{-1}\mathbf{p}_1-\check{\gamma}D_{j}^\text{DA}\mathbf{K}^{-1}\mathbf{p}+\check{z}\mathbf{K}^{-1}(\mathbf{p}_1-\mathbf{p}),
\end{equation}
where $D_{j,1}^\text{DA}=\mathbf{D}_j^\text{DA}(\mathbf{p}_1)$.
\par Since $\mathbf{p}_1$ and $\mathbf{p}$ are neighborhood pixel and significantly close, the offset vector can further be approximated as,
\begin{equation}
    \mathbf{v}_1\approx\check{\gamma}(D_{j,1}^\text{DA}\mathbf{K}^{-1}\mathbf{p}_1-D_{j}^\text{DA}\mathbf{K}^{-1}\mathbf{p}),
\end{equation}
where $\mathbf{v}_1$ can again be expressed as a vector scaled by a constant factor $\check{\gamma}$, analogous to the derivation presented in the main manuscript using the estimated depth map in the original target view.
\par In this manner, we eliminate the influence of scale and shift introduced by spatial dense depth reconstruction of DepthAnything V2, enabling the formulation of spatial surface normal vector map in target view and the computation of surface normal consistency loss $\tilde{\mathcal{L}}_{\text{SNC}}$.
\par Notably, as scale-ambiguous depth cannot be used for spatial warping to construct photometric losses or any direct supervision signal, this spatial normal vector map derived from reconstructed dense depth can neither be obtained via forward warping, as pointing direction of elements are not coherent across view; nor via backward warping or modified forward-backward schemes, since camera extrinsics are calibrated in the metric physical world, whereas the depth output of DepthAnything V2 remains scale-ambiguous.

\subsection{Motion learning methods}
\par In this work, we propose an adaptive joint motion learning strategy. Here, we present a detailed formulation of how prior works have realized surround-view motion learning.
\par FSM \cite{fsm} estimates the pose of each camera independently. This motion learning paradigm can be written as,
\begin{equation}
    \{\boldsymbol{f}_i\}_{i=1}^N=\{\mathcal{P}_\text{en}(\mathbf{I}_i^t,\mathbf{I}_i^{t'})\}_{i=1}^N,
\end{equation}
\begin{equation}
    \{\hat{\mathbf{T}}_i^{t\rightarrow t'}\}_{i=1}^N=\{\mathcal{P}_\text{de}(\boldsymbol{f}_i)\}_{i=1}^N,
\end{equation}
where $\mathcal{P}_\text{en}$ and $\mathcal{P}_\text{de}$ indicates en- and decoder of pose network. 
\par In addition, it enforces pose consistency by transforming all pose estimation to the coordinate of front camera as a global constraint. The pose transformation can be written as,
\begin{equation}
    \tilde{\mathbf{T}}_i^{t\rightarrow t'}=\mathbf{E}_1^{-1}\mathbf{E}_i\hat{\mathbf{T}}_i^{t\rightarrow t'}\mathbf{E}_i^{-1}\mathbf{E}_1
\end{equation}
where $\mathbf{E}_1$ indicates extrinsics of front camera. $\tilde{\mathbf{T}}_i^{t\rightarrow t'}=[\tilde{\mathbf{R}}_i^{t\rightarrow t'},\tilde{\mathbf{t}}_i^{t\rightarrow t'}]$. Subsequently, it formulated pose consistency on translation and rotation separately as,
\begin{equation}
    \mathbf{t}_\text{loss}=\sum_{j=2}^N\Vert\hat{\mathbf{t}}_1^{t\rightarrow t'}-\tilde{\mathbf{t}}_j^{t\rightarrow t'}\Vert^2,
\end{equation}
\begin{equation}
    \mathbf{R}_\text{loss}=\sum_{\varrho\in\{\phi,\theta,\psi\} }\sum_{j=2}^N\Vert\hat{\varrho}_1^{t\rightarrow t'}-\tilde{\varrho}_j^{t\rightarrow t'}\Vert^2,
\end{equation}
\begin{equation}
    \mathcal{L}_\text{PCC}=\alpha_t\mathbf{t}_\text{loss}+\alpha_r\mathbf{R}_\text{loss},
\end{equation}
where $\alpha_t$ and $\alpha_r$ are weighting coefficients.
\par SurroundDepth \cite{surrounddepth} proposes a joint motion estimation strategy that aggregates feature maps from all cameras using a shared pose encoder and estimates a unified ego-motion in the LiDAR coordinate frame via a pose decoder. The pose of each individual camera is then recovered by distributing the joint motion through the calibrated extrinsic parameters.
\begin{equation}
    \hat{\mathbf{T}}_i^{t\rightarrow t'}=\mathbf{E}_i^{-1}\mathcal{P}_\text{de}\left(\frac{1}{N}\sum_{i=1}^N\boldsymbol{f}_i\right)\mathbf{E}_i,
\end{equation}
\par Both VFDepth \cite{vfdepth} and CVCDepth \cite{cvcdepth} focus on estimating the front-camera motion. VFDepth conditions the pose decoder on aggregated multi-camera features, while CVCDepth relies solely on features from the front camera. The resulting motion estimation is formulated as,
\begin{equation}
    \hat{\mathbf{T}}_i^{t\rightarrow t'}=\mathbf{E}_i^{-1}\mathbf{E}_1\mathcal{P}_\text{de}(\boldsymbol{f}_1)\mathbf{E}_1^{-1}\mathbf{E}_i,
\end{equation}
\par The alternative of motion learning adopted by CVCDepth \cite{cvcdepth} can be intuitively attributed to the observation that spatial structure within FoV of front camera are, in most driving scenarios, more sensitive to ego-motion and therefore provide informative cues for SfM learning. However, this heuristic implicitly assumes a fixed dominance of the front view and overlooks the complementary motion cues available from other camera views. Therefore, in this work, we propose a adaptive joint motion learning strategy, in which the pose network adaptively learns and weights the contribution of each camera view for motion estimation.

\section{Experiments}
\label{sec::experiment}
\subsection{Implementation details}
\par The weighting coefficients of the loss functions were set as $\lambda_\text{T}=1$, $\lambda_\text{S}=0.03$, $\lambda_\text{ST}=0.1$, $\lambda_\text{MVRC}=0.2$, $\omega_p=1$, $\omega_s=0.001$, $\omega_\text{SDC}=0.001$, $\omega_\text{SNC}=0.01$, $\omega_\text{DSC}=1$, $\kappa_\text{SRC}=0.1$, $\kappa_\text{SNC}=0.1$, $\mu=0.1$. 
\par ViT-Base variant of DepthAnything V2 and ViT-B/32 variant of CLIP model were used as foundation models. Self-occlusion masks and reprojection masks were applied to exclude invalid pixels from loss computation.

\subsection{Evaluation metrics} 
\par The evaluation metrics used for our experiments are calculated as follows. 
\begin{itemize}
    \item Absolute relative error (Abs Rel):
    \[
        \frac{1}{n}\sum_{i\in n}|\hat{\mathbf{D}}(i)-\mathbf{D}(i)|/\mathbf{D}(i);
    \]
    \item Squared relative difference (Sq Rel): 
    \[
        \frac{1}{n}\sum_{i\in n}\Vert\hat{\mathbf{D}}(i)-\mathbf{D}(i)\Vert^2/\mathbf{D}(i);
    \]
    \item Root mean squared error (RMSE): 
    \[
        \sqrt{\frac{1}{n}\sum_{i\in n}\Vert\hat{\mathbf{D}}(i)-\mathbf{D}(i)\Vert^2};
    \]
    \item Root mean squared logarithmic error (RMSE log):
    \[
        \sqrt{\frac{1}{n}\sum_{i \in n}\Vert\log\hat{\mathbf{D}}(i)-\log\mathbf{D}(i)\Vert^2}
    \]
    \item Accuracy with threshold ($\delta_t$):
    \[
        \%~\text{of}~\hat{\mathbf{D}}(i)~\text{s.t.}~ \max\left(\dfrac{\hat{\mathbf{D}}(i)}{\mathbf{D}(i)},\dfrac{\mathbf{D}(i)}{\hat{\mathbf{D}}(i)}\right)<1.25^t,
    \]
\end{itemize}
where $n$ indicates number of valid depths in groundtruth.

\subsection{Experiment results on KITTI}
\par In this section, we present detailed experimental results for scale-ambiguous depth estimation on the KITTI Eigen raw split. For fair comparison, we consider only methods that 
perform inference using a single input image. We denote our method as \textbf{GeoDepth} to reflect its self-supervised monocular depth estimation implementation. Concretely, the network architecture remains the same as that of GeoSurDepth, while the loss function was modified as,
\begin{equation}
    \mathcal{L}=\omega_p\mathcal{L}_p+\omega_s\mathcal{L}_s+\omega_\text{SNC}\mathcal{L}_\text{SNC}+\omega_\text{DSC}\mathcal{L}_\text{DSC},
\end{equation}
where photometric loss $\mathcal{L}_p$ only contains temporal context. $\omega_p=1$, $\omega_s=0.001$, $\omega_\text{SNC}=0.01$, $\omega_\text{DSC}=1$. All other hyperparameters follow the settings of MonoViT.
\par The quantitative results reported in Table \ref{table::kitti_all} demonstrate the SoTA performance of our method compared with other baselines. Notably, this performance is achieved efficiently and with minimal additional complexity, by simply modifying the network architecture and incorporating additional loss terms, without introducing auxiliary modalities such as semantic segmentation, optical flows, or temporal cues.
\begin{table*}[h]
\renewcommand{\arraystretch}{1}
\centering
\scalebox{0.95}{
\begin{tabular}{c|*{7}{c}}
\toprule
Method & \cellcolor{red!8}{Abs Rel$\downarrow$} & \cellcolor{red!8}{Sq Rel$\downarrow$} & \cellcolor{red!8}{RMSE$\downarrow$} & \cellcolor{red!8}{RMSE$_\text{log}$$\downarrow$} & \cellcolor{cyan!8}{$\delta_1$$\uparrow$} & \cellcolor{cyan!8}{$\delta_2$$\uparrow$} & \cellcolor{cyan!8}{$\delta_3$$\uparrow$}\\
\midrule
Monodepth2 \cite{godard2019digging} & 0.115 & 0.903 & 4.863 & 0.193 & 0.877 & 0.959 & 0.981 \\
SAFENet \cite{choi2020safenet} & 0.112 & 0.788 & 4.582 & 0.187 & 0.878 & 0.963 & 0.983 \\
PackNet-SfM \cite{guizilini20203d} & 0.111 & 0.785 & 4.601 & 0.189 & 0.878 & 0.960 & 0.982\\
HR-Depth \cite{lyu2021hr} & 0.109 & 0.792 & 4.632 & 0.185 & 0.884 & 0.962 & 0.983\\
FSRE-Depth \cite{jung2021fine} & 0.102 & 0.675 & 4.393 & 0.178 & 0.893 & 0.966 & 0.984\\
BRNet \cite{han2022brnet} & 0.105 & 0.698 & 4.462 & 0.179 & 0.890 & 0.965 & 0.984\\
SD-SSMDE \cite{petrovai2022exploiting} & 0.100 & 0.661 & 4.264 & 0.172 & 0.896 & 0.967 & \underline{0.985} \\
RA-Depth \cite{he2022ra} & 0.096 & \underline{0.632} & 4.216 & 0.171 & 0.903 & 0.968 & \underline{0.985}\\
MonoFormer \cite{bae2023deep} & 0.104 & 0.846 & 4.580 & 0.183 & 0.891 & 0.962 & 0.982\\
GPGS-Depth (CC) \cite{jia2022self} & 0.116 & 0.842 & 4.708 & 0.190 & 0.876 & 0.961 & 0.982\\
freq-aware-depth \cite{chen2023frequency} & 0.105 & 0.711 & 4.452 & 0.181 & 0.886 & 0.964 & 0.984\\
Lite-Mono \cite{zhang2023lite} & 0.101 & 0.729 & 4.454 & 0.178 & 0.897 & 0.965 & 0.983\\
DaCCN \cite{han2023self} & 0.099 & 0.661 & 4.316 & 0.173 & 0.897 & 0.967 & \underline{0.985}\\
DNA-Depth \cite{shen2023dna} & 0.102 & 0.757 & 4.493 & 0.178 & 0.896 & 0.965 & 0.984\\
EDS-Depth \cite{yu2025eds} & \underline{0.095} & \textbf{0.619} & \underline{4.184} & 0.170 & \underline{0.905} & \underline{0.969} & \underline{0.985}\\
MonoViT (MPViT-tiny) \cite{monovit} & 0.102 & 0.733 & 4.459 & 0.177 & 0.895 & 0.965 & 0.984\\
MonoViT (MPViT-small) \cite{monovit} & 0.099 & 0.708 & 4.372 & 0.175 & 0.900 & 0.967 & 0.984\\
MonoViT (MPViT-base) \cite{monovit} & 0.100 & 0.747 & 4.427 & 0.176 & 0.901 & 0.966 & 0.984\\
\textbf{GeoDepth} (MPViT-tiny) & 0.098 & 0.661 & 4.230 & 0.170 & 0.900 & 0.968 & \textbf{0.986}\\
\textbf{GeoDepth} (MPViT-small) & 0.096 & 0.649 & 4.199 & \underline{0.168} & 0.903 & \underline{0.969} & \textbf{0.986}\\
\textbf{GeoDepth} (MPViT-base) & \textbf{0.094} & 0.641 & \textbf{4.175} & \textbf{0.167} & \textbf{0.906} & \textbf{0.970} & \textbf{0.986}\\
\bottomrule
\end{tabular}
}
\caption{Scale-ambiguous evaluation on KITTI eigen test split. All methods are evaluated using the raw instead of improved depth groundtruth and without horizontal-flip post-processing. Resolution of input image to the depth network was $(192,640)$. 
}
\label{table::kitti_all}
\end{table*}

\subsection{Cross-dataset generalization evaluation}
\par
\begin{figure}[h]
    \centering
    \includegraphics[width=1\linewidth]{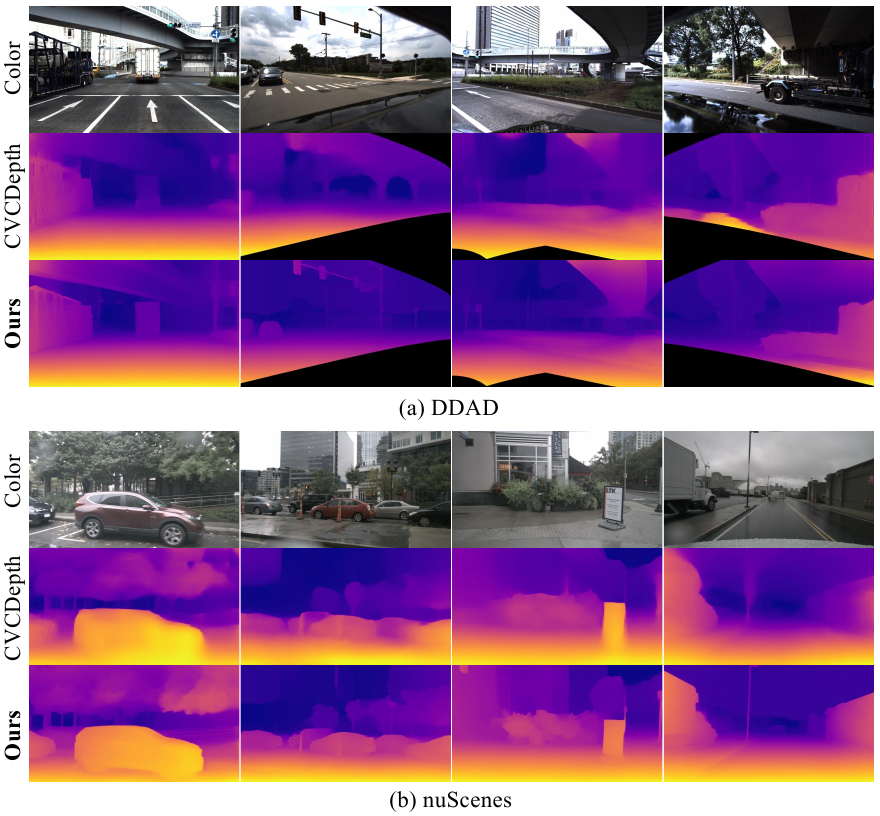}
    \caption{Visualization of cross-dataset evaluation on both datasets.}
    \label{fig::cross}
\end{figure}

 In Fig.\ref{fig::cross}, we present qualitative visualizations of the cross-dataset evaluation, comparing our proposed method with CVCDepth, in addition to the quantitative results reported in the main manuscript. The results demonstrate the superior generalization capability of our method. In particular, when the model is trained on the nuScenes dataset and directly evaluated on DDAD, it exhibits strong generalization performance that is even competitive with methods trained directly on DDAD. This can be attributed to the fact that nuScenes is a more challenging benchmark, containing diverse and adverse conditions such as rainy scenes, enabling the model to generalize well to comparatively simpler datasets like DDAD.

\subsection{Addition visualization of experiment results}
\par
\begin{figure*}[t]
    \centering
    \includegraphics[width=1\linewidth]{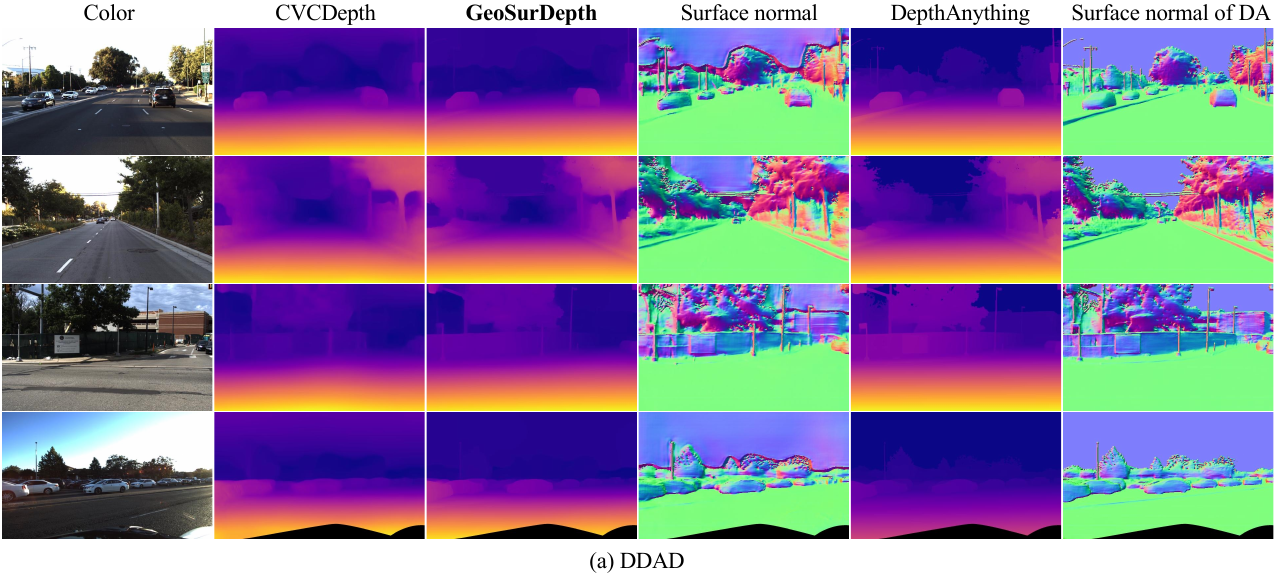}
    \includegraphics[width=1\linewidth]{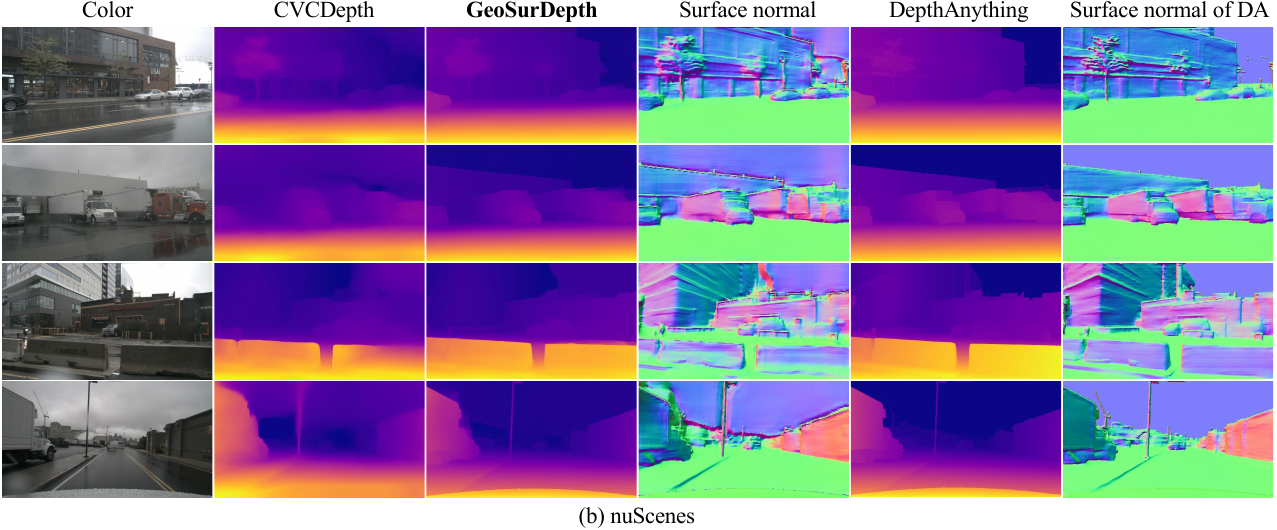}
    \caption{More comparison examples on the DDAD and nuScenes datasets are presented. Our method accurately estimates edge-aware, naturally transitioning, and smooth depth under diverse conditions.}
    \label{fig::disp_extra}
\end{figure*}

 In Fig.\ref{fig::disp_extra}, we present additional examples of depth estimation, surface normal visualization, as well as pseudo-depth of DepthAnything V2 and the corresponding surface normal maps computed from it on the DDAD and nuScenes datasets. These results further validate the effectiveness of our method in producing edge-aware, naturally transitioning, and smooth depth estimates under diverse conditions. 
\par
\begin{figure*}[t]
    \centering
    \includegraphics[width=1\linewidth]{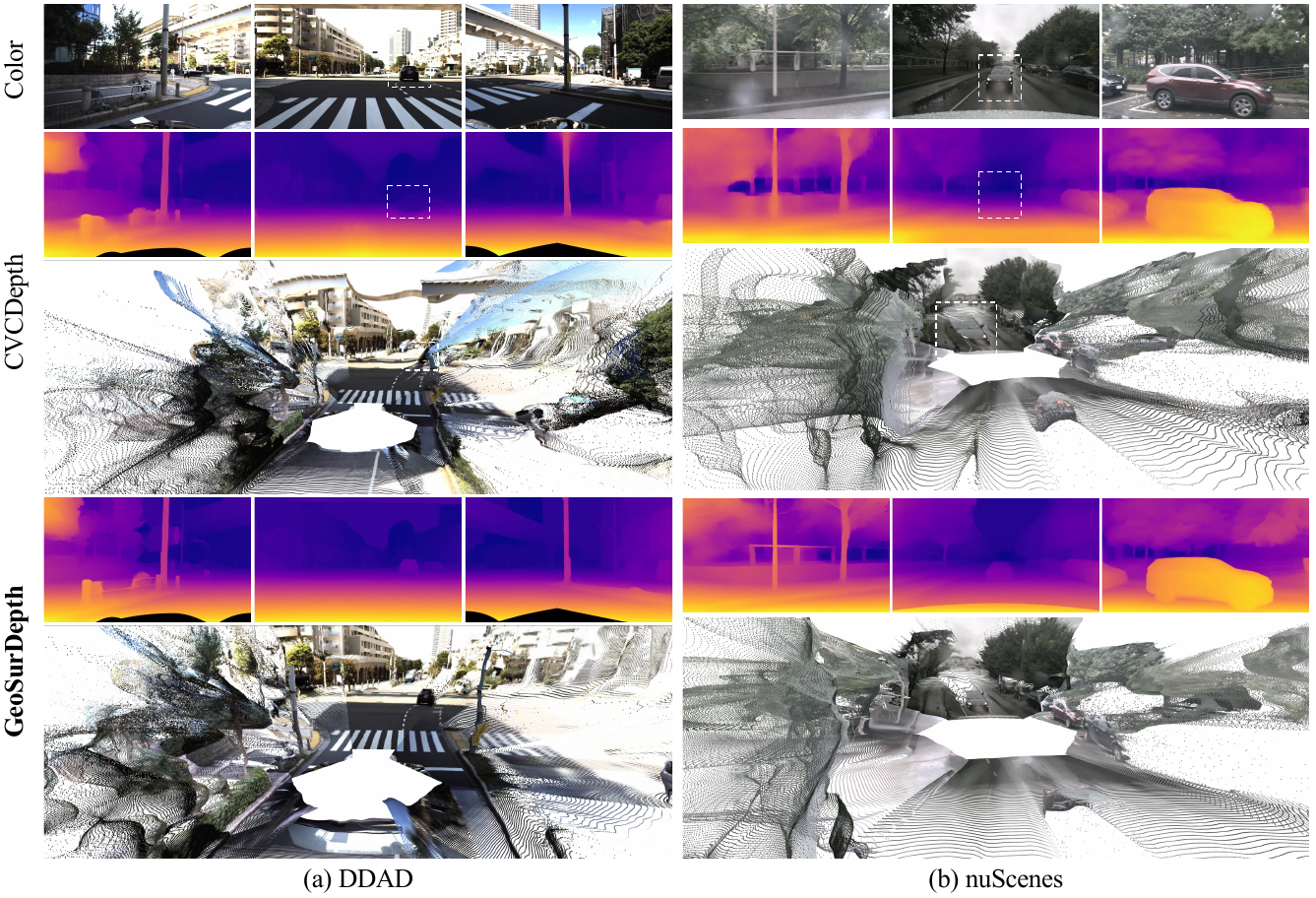}
    \caption{Examples of point cloud reconstruction comparison on DDAD and nuScenes datasets.}
    \label{fig::cloud_extra}
\end{figure*}

 In Fig.\ref{fig::cloud_extra}, we present examples of point cloud reconstruction using estimated dense depth on the DDAD and nuScenes datasets, with comparisons against the baseline method CVCDepth \cite{cvcdepth}. The visualizations show that our proposed method produces geometrically regular point clouds with improved cross-view coherence and spatial consistency. For example, in Fig.\ref{fig::cloud_extra}(a), lane markings are cleanly aligned across views, road lights stand upright above the ground plane, and distant vehicles are accurately projected and positioned. In contrast, in Fig.\ref{fig::cloud_extra}(b), the baseline method incorrectly estimates the depth of a vehicle in the front view, resulting in erroneous 3D projection, while depth holes in the rear view further cause the vehicle to be wrongly projected onto the ground.

\subsection{Ablation study on DepthAnything V2 encoder}
\par We conduct experiments using different encoder variants of DepthAnything V2. As shown in Table \ref{table::ab_encoder}, the different encoders exhibit largely comparable performance, with ViT-B achieving slightly better results. We attribute this to the relatively low input image resolution used by the network, under which fine-grained details may be compressed, limiting the advantage of more powerful encoders.
\begin{table}[h]
\renewcommand{\arraystretch}{1}
\setlength{\tabcolsep}{3.3pt}
\centering
\scalebox{0.95}{
\begin{tabular}{c|*{7}{c}}
\toprule
Encoder & \cellcolor{red!8}{Abs Rel$\downarrow$} & \cellcolor{red!8}{Sq Rel$\downarrow$} & \cellcolor{red!8}{RMSE$\downarrow$} & \cellcolor{red!8}{RMSE$_\text{log}$$\downarrow$} & \cellcolor{cyan!8}{$\delta_1$$\uparrow$} & \cellcolor{cyan!8}{$\delta_2$$\uparrow$} & \cellcolor{cyan!8}{$\delta_3$$\uparrow$}\\
\midrule
ViT-S & 0.177 & 2.782 & \underline{11.639} & \underline{0.284} & \underline{0.759} & \underline{0.909} & \underline{0.955}\\
\textbf{ViT-B} & \underline{0.176} & \underline{2.738} & \textbf{11.520} & \textbf{0.280} & \textbf{0.763} & \textbf{0.912} & \textbf{0.957}\\
ViT-L & \textbf{0.175} & \textbf{2.682} & 11.895 & \underline{0.284} & 0.754 & 0.906 & \underline{0.955}\\
\bottomrule
\end{tabular}
}
\caption{Comparison of metric depth estimation results with different encoder of DepthAnything V2 on DDAD dataset.}
\label{table::ab_encoder}
\end{table}

\section{Code}
Our framework is built upon the open-source implementations of CVCDepth \cite{cvcdepth}, MonoViT \cite{monovit}, VFDepth \cite{vfdepth}, and Monodepth2 \cite{godard2019digging}. In the supplementary material, we provide part of our implementation scripts related to the key loss functions and surface normal vector map computation for your reference.

\bibliographystyle{IEEEtran}
\bibliography{reference.bib}